%% file: top.tex
\documentclass[runningheads]{llncs}
\usepackage{graphicx}
\usepackage{comment}
\usepackage{amsmath,amssymb} % define this before the line numbering.
\usepackage{color}
\usepackage{epsfig}
\usepackage{graphicx}
\usepackage{xspace}
\usepackage{color}
\usepackage{xcolor}
\usepackage{xspace}
\usepackage{multirow}
\usepackage[caption=false]{subfig}
\usepackage{tabulary}
\usepackage{pifont}

\usepackage[width=132mm,left=15mm,paperwidth=162mm,height=193mm,top=12mm,paperheight=217mm]{geometry}

\makeatletter
\DeclareRobustCommand\onedot{\futurelet\@let@token\@onedot}
\def\@onedot{\ifx\@let@token.\else.\null\fi\xspace}
\def\eg{\emph{e.g}\onedot} 
\def\ie{\emph{i.e}\onedot}

\makeatother

\begin{document}
\pagestyle{headings}
\mainmatter
\def\ECCVSubNumber{}  % Insert your submission number here

\title{Learning Lane Graph Representations \\for Motion Forecasting}
\titlerunning{Learning Lane Graph Representations for Motion Forecasting}

\author{Ming Liang\inst{1} \and Bin Yang\inst{1,2} \and Rui Hu\inst{1} \and Yun Chen\inst{1} \and Renjie Liao\inst{1,2} \and Song Feng\inst{1} \and Raquel Urtasun\inst{1,2}}
\authorrunning{M. Liang et al.}
\institute{Uber ATG \and University of Toronto\\
\email{\{ming.liang,byang10,rui.hu,yun.chen,rjliao,songf,urtasun\}@uber.com}}
\maketitle

\input{abstract}
\input{introduction}
\input{related}

\input{model}
\input{experiments}
\input{conclusion}
\input{acknowledgement}
\input{appendix}

\bibliographystyle{splncs04}
\bibliography{egbib}
\end{document}

%% file: abstract.tex
\begin{abstract}
We propose a motion forecasting model that exploits a novel structured map representation as well as actor-map interactions. 
Instead of encoding vectorized maps as raster images, we construct a lane graph from raw map data to explicitly preserve the map structure.
To capture the complex topology and long range dependencies of the lane graph, we propose LaneGCN which extends graph convolutions with multiple adjacency matrices and along-lane dilation. To capture the complex interactions between actors and maps, we exploit a fusion network consisting of four types of interactions, actor-to-lane, lane-to-lane, lane-to-actor and actor-to-actor.
Powered by LaneGCN and actor-map interactions, our model is able to predict accurate and realistic multi-modal trajectories. 
Our approach significantly outperforms the  state-of-the-art on the large scale Argoverse motion forecasting benchmark.
	\keywords{HD Map, Motion Forecasting, Autonomous Driving.}
\end{abstract}

%% file: introduction.tex
\section{Introduction}
\begin{figure*}[t]
	\begin{center}
		\includegraphics[width=1.0\linewidth]{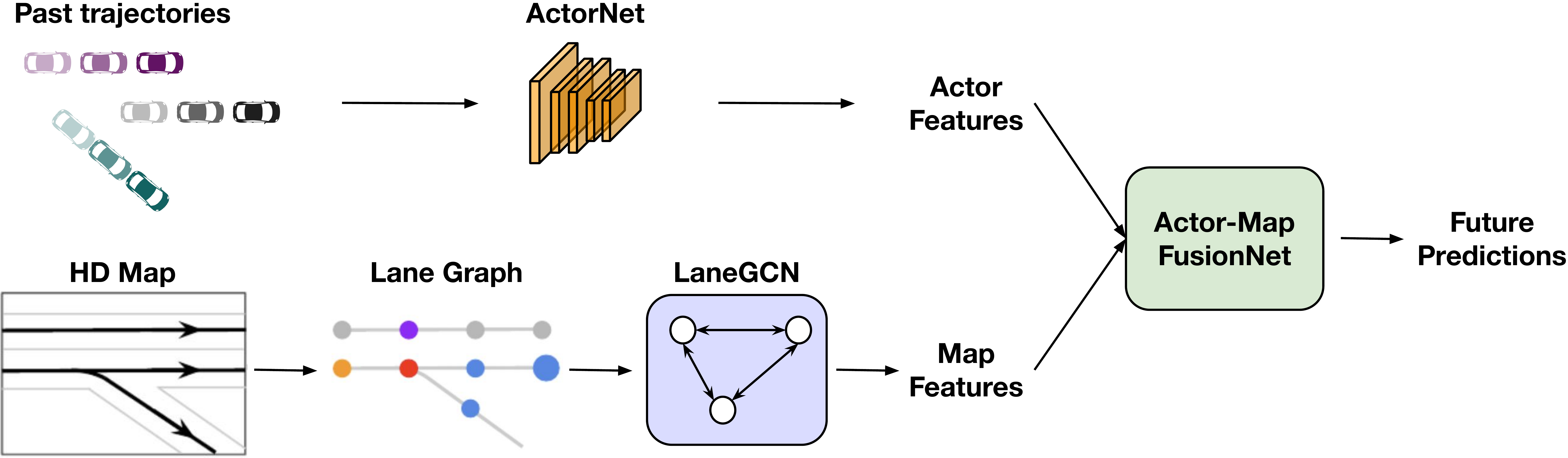}
	\end{center}
	\caption{\textbf{Our approach:} We construct a lane graph from raw map data and use   LaneGCN to extract map features. In parallel, ActorNet extracts actor features from observed past trajectories. We then use FusionNet to model the Interactions between actors themselves and the map, and predict the  future trajectories.}
	\label{fig:concept}
\end{figure*}

Autonomous driving has the potential to revolutionize   transportation.  Self-driving  vehicles (SDVs) 
have to accurately predict the future motions of other traffic participants in order 
to safely operate.
High Definition maps (HD-maps) provide extremely useful geometric and semantic information for motion forecasting, as the behaviors of actors largely depend on the map topology. For example, a vehicle is unlikely to take a left turn when there is not a left turn lane nearby. Effectively exploiting HD maps is essential for motion forecasting models to produce plausible and accurate trajectories.

First attempts exploit HD maps as heuristics \cite{bertha2014}. Actors are first associated with lanes and all candidate motion paths are then generated based on map topology. In this way, the prediction results are constrained by the map. However, this approach can not capture rare and non-compliant behaviours, which while not very likely, might be safety critical. 

Recent works \cite{hdnet,djuric2018motion,pnpnet,chauffeurnet,hong2019rules,intentNet,ilvm,prior} use machine learning  to learn semantic representations from maps. To enable HD maps to be processed by neural networks the map data is rasterized to create image-like raster inputs. Map topology is implicitly encoded as lines, masks or colours, which are then processed by a 2D Convolutional Neural Network (CNN). These learned map features were shown to provide useful context information for motion forecasting.
However, these approach has two disadvantages. First, the rasterization process inevitably results in  information loss. Second, maps have a graph structure with complex topology which 2D convolution may be very inefficient to capture. For example, a lane of interest may extend for a long range in the lane direction. To capture this information, the receptive field has to be very large, covering not only the intended area, but also large areas outside the lane. 
Furthermore, lane pairs in the same  or opposite directions have completely different semantic meanings and dependencies, although the lanes in both pairs are spatially close to each other.

 In this paper we made three  main contributions: (1) Instead of using rasterization, we construct a lane graph from vectorized map data, thus avoiding information loss. We then propose the Lane Graph Convolutional Network (LaneGCN), which  effectively captures the complex topology and long range dependencies of the lane graph.
(2) Based on  LaneGCN, our motion forecasting model  captures all possible actor-map interactions. 
In particular, we represent both actors and lanes  as nodes in the graph and  use a 1D CNN and  LaneGCN to extract the features for the actor and lane nodes respectively, and then exploit spatial attention and another LaneGCN to model four types of interactions: {\it actor-to-lane}, {\it lane-to-lane}, {\it lane-to-actor} and {\it actor-to-actor}.
We refer the reader to Fig. \ref{fig:concept} for  an illustration of our approach.
%Equipped with  LaneGCN and actor-map and actor-actor interactions, our model is able to predict plausible and accurate multi-modal trajectories.
(3) We conduct experiments on the large-scale Argoverse  motion forecasting benchmark \cite{chang2019argoverse}, and show  significant improvements over the  state-of-the-art.

%% file: related.tex
\section{Related Work}\label{sec:related}
In this section, we review  work on  map representations, learning map representations for autonomy tasks, and graph convolutional networks.

\noindent{\bf Map Representations:}
HD  maps capture both the lane geometry as well as their connectivity.
\cite{Homayounfar2018HierarchicalRA} proposes to
parameterize the lane boundaries as a set of polylines, and exploit a Recurrent Neural Network (RNN) to extract them from sensor data. \cite{Liang2019ConvolutionalRN} further extends the polyline representation to a more structured parameterization. 
Instead of modelling the geometry of each lane, \cite{HomayounfarDAGMapperLT} proposes to
parameterize the unknown lane graph  as a Directed Acyclic Graphical model (DAG), which is more robust and able to handle more complex topology like branching. In addition to modelling the geometry, \cite{Mttyus2016HDMF,Mttyus2015EnhancingRM} encode different lane types in a graphical model to better exploit their appearance features. \cite{Chu2019NeuralTG} parameterizes the  road layout using an undirected graph,  showcasing outstanding performance in  large-scale city scale road topology.

\noindent{\bf Learning Map Representations for Autonomy:}
Rasterization based map representations have been extensively used.
\cite{djuric2018motion,Cui2018MultimodalTP,Chou2019PredictingMO} rasterize map elements (roads, crosswalks) as layers and encode the lane direction with different colors.
\cite{chauffeurnet,Chai2019MultiPathMP} encode roadmap, traffic lights and speed limits in rasterized bird's eye view images.
\cite{hong2019rules} encodes the history of static entities, dynamic entities and semantic map information in a top-down spatial grid.
HDNet \cite{hdnet} exploits the road mask as input feature to improve object detection performance.
Rasterized maps have been fused with  LiDAR point clouds to perform joint perception and prediction \cite{pnpnet,spagnn,interacttransform2020}  as well as  end-to-end motion  planning \cite{nmp,ppp,dsd}.
While raster map representations are popular, an alternative is to use vectorized map features. \cite{chang2019argoverse} uses the distance along the centerlines and offset from the centerlines as input to their nearest neighbours regression and LSTM \cite{hochreiter1997long} models.  \cite{jean2019,argoai_challenge} use 1D CNN and LSTM to encode lane features. In contrast, our model constructs a lane graph from  vectorized map data, and extracts multi-scale topology features using the proposed LaneGCN.
In concurrent work VectorNet\cite{vectornet},  two graph networks are used  to extract actor/lane features and model global interactions, respectively. 
There are two major differences between VectorNet and LaneGCN. First, VectorNet uses vanilla graph networks with undirected full connections, while we build a sparsely connected lane graph following the map topology and propose task specific multi-type and dilated graph operators. Second, VectorNet uses polyline-level nodes for interaction, while our LaneGCN uses polyline segments as map nodes to capture higher resolution. Note that in our approach nodes in different polylines can interact with each other through dilated connections.

\noindent{\bf Graph Convolutional Networks:}
Graph Convolutional Networks (GCNs) \cite{shuman2013emerging,henaff2015deep,duvenaud2015convolutional,kipf2016semi,defferrard2016convolutional,liao2019lanczosnet} have been shown to be effective for graph representation learning.
They generalize the 2D convolution on grids to arbitrary graphs via the so called graph convolution. Different from 2D convolution, which operates on neighbors in a local grid, graph convolution operates on the neighboring nodes defined by the graph structure, typically described in the form of an adjacency matrix.
We draw inspiration from GCNs and propose  LaneGCN, which is a specialized version designed for lane graphs. In our model, we introduce multiple adjacency matrices and multi-scale dilated convolutions, which are effective in capturing the complex topology and long-range dependencies of the lane graph.

%% file: model.tex
\section{Lane Graph Representations for Motion Forecasting}\label{sec:model}

\begin{figure*}[t]
	\begin{center}
		\includegraphics[width=1.0\linewidth]{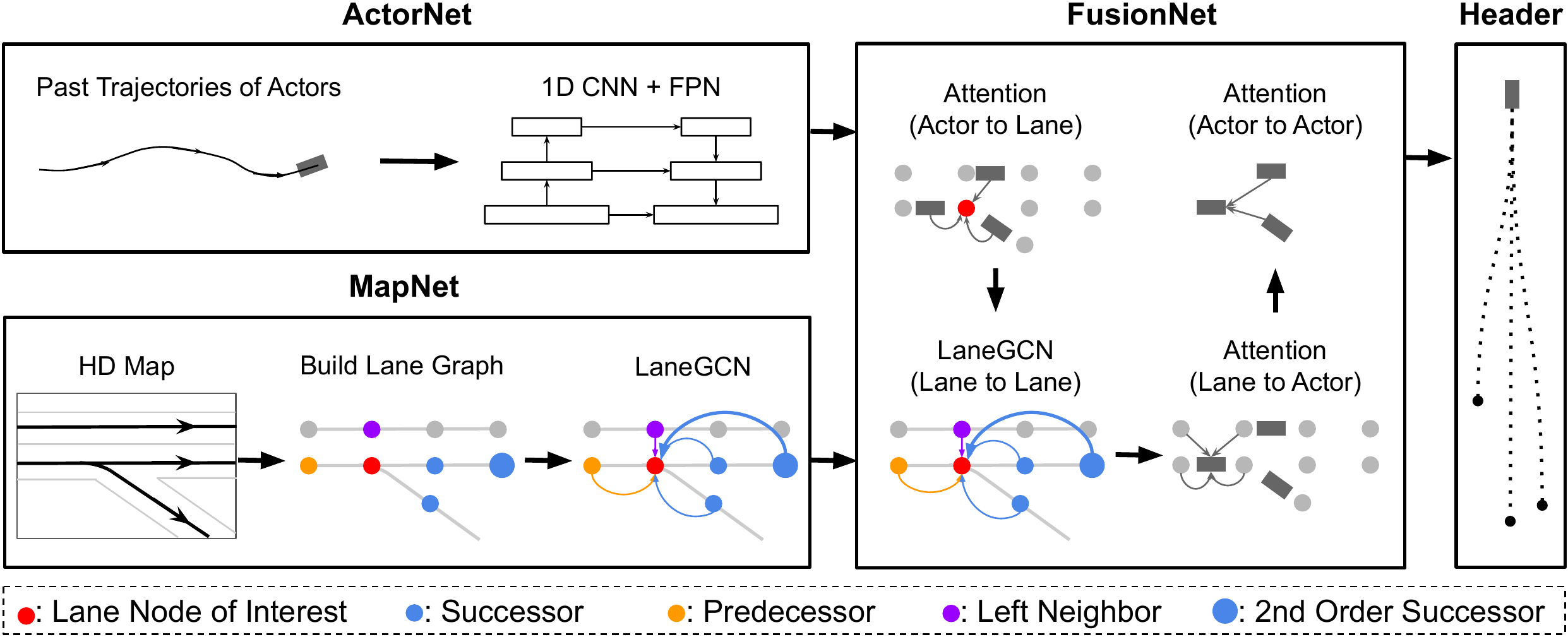}
	\end{center}
	\caption{\textbf{Overall architecture:}
	Our model is composed of four modules.
	\textbf{(1)} ActorNet receives the past actor trajectories as input, and uses 1D convolution to extract actor node features.
	\textbf{(2)} MapNet constructs a lane graph from HD maps, and uses a LaneGCN to exact lane node features.
	\textbf{(3)} FusionNet is a stack of 4 interaction blocks. The actor to lane block fuses real-time traffic information from actor nodes to lane nodes. The lane to lane block propagates information over the lane graph and updates lane features. The lane to actor block fuses updated map information from lane nodes to actor nodes. The actor to actor block performs interactions among actors. We use another LaneGCN for the lane to lane block, and spatial attention layers for the other blocks.
	\textbf{(4)} The prediction header uses after-fusion actor features to produce multi-modal trajectories.  
	}
	\label{fig:overall}
\end{figure*}

In this section, we propose a novel motion forecasting model that learns structured map representations and  fuses the information of traffic actors and HD maps taking into account their interactions.
In the following, we  explain the four modules that compose our model, \ie,  how to compute actor features with \textbf{ActorNet}, how to represent the map via \textbf{MapNet}, how to fuse the information from both actors and the map with  \textbf{FusionNet}, and finally how to predict the final motion forecasting trajectories through the \textbf{Prediction Header}.
We refer the reader to Fig. \ref{fig:overall} for an illustration of the overall architecture.

\subsection{ActorNet: Extracting Traffic Participant Representations}\label{sec:actor}
We assume actor data is composed of the observed past trajectories of all actors in the scene.
Each trajectory is represented as a sequence of displacements $\{ \Delta\mathbf{p}_{-(T-1)}, \dots, \Delta\mathbf{p}_{-1}, \Delta\mathbf{p}_0 \}$, where $\Delta\mathbf{p}_t$ is the 2D displacement from time step $t-1$ to $t$, and $T$ is the trajectory size. 
All coordinates are defined in the Bird's Eye View (BEV), as this is the space of interest for traffic agents.
For trajectories with sizes  smaller than $T$, we pad them with zeros. 
We add a binary $1\times T$ mask to indicate if the element at each step is padded or not and concatenate it with the trajectory tensor, resulting in an input tensor of size $3\times T$. 

While both CNNs and RNNs can be used for temporal data, here we use an 1D CNN to process the trajectory input for its effectiveness in extracting multi-scale features and efficiency in parallel computing. The output of ActorNet is a temporal feature map, whose element at $t=0$ is used as the actor feature.
The network has $3$ groups/scales of 1D convolutions. 
Each group consists of $2$ residual blocks \cite{residual}, with the stride of the first block as $2$. 
We then use a Feature Pyramid Network (FPN) \cite{Lin2016FeaturePN} to fuse the multi-scale features, and apply another residual block to obtain the output tensor. For all layers, the convolution kernel size is $3$ and the number of output channels is $128$. Layer normalization \cite{layernorm} and  the Rectified Linear Unit (ReLU) \cite{glorot2011deep} are used after each convolution.

\subsection{MapNet: Extracting Structured Map Representation}\label{sec:map}
We use a novel deep model, called MapNet, to learn structured map representations from vectorized map data. This contrasts previous approaches, which encode the map as a raster image and apply 2D convolutions to extract features. 
MapNet consists of two steps: (1) building a lane graph from vectorized map data; (2) applying our  novel LaneGCN to the lane graph to output the map features.

\subsubsection{Map Data:}
\begin{figure*}[t]                                                                               
	\begin{center}
		\includegraphics[width=0.85\linewidth]{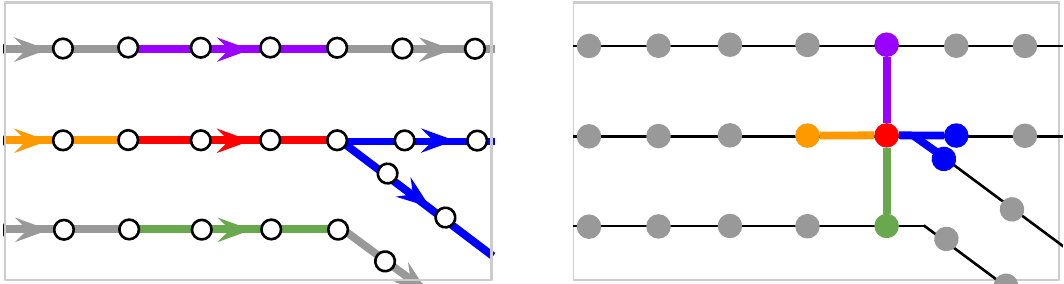}
	\end{center}
	\caption{\textbf{Lane graph construction from vectorized map data}. \textbf{Left}: 
		The lane centerline of interest, its predecessor, successor, left and right neighbor are denoted with red, orange, blue, purple, and green lines, respectively. Each centerline is given as a sequence of BEV points (hollow circles). \textbf{Right}: Derived lane graph with an example lane node. The lane node of interest, its predecessor, successor, left and right neighbor are denoted with red, orange, blue, purple and green circles respectively. See Section \ref{sec:construct_graph} for more information.}
	\label{fig:data}
\end{figure*}

In this paper, we adopt a simple form of vectorized map data as our representation of HD maps.
Specifically, the map data is represented as a set of lanes and their connectivity. 
Each lane contains a centerline, \ie, a sequence of 2D BEV points, which are arranged following the lane direction (see Fig. \ref{fig:data}, top).
For any two lanes which are directly reachable, $4$ types of connections are given: \textit{predecessor}, \textit{successor}, \textit{left neighbour} and \textit{right neighbour}. 
Given a lane $A$, its predecessor and successor are the lanes which can directly travel to $A$ and from $A$ respectively. 
Left and right neighbours refer to the  lanes which can be directly reached without violating traffic rules.
This simple map format provides essential geometric and semantic information for motion forecasting, as vehicles  generally plan their routes by reference to lane centerlines and their connectivity. 

\subsubsection{Lane Graph Construction:} \label{sec:construct_graph}
Instead of encoding maps as raster images, we derive a lane graph from the map data as the input. In designing the lane graph, we expect its nodes to have a fine resolution. Given any actor location, we query the lane graph and find its nearest nodes to retrieve accurate map information. From this point of view, it is not an optimal choice to directly use the lane centerlines as the nodes.

We refer the reader to Fig.  \ref{fig:data} for an example of the lane graph construction.
We first define a lane node as the straight line segment formed by any two consecutive points (grey circles in Fig. \ref{fig:data}) of the centerline.
The location of a lane node is the averaged coordinates of its two end points.
Following the  connections between lane centerlines, we also derive $4$ connectivity types for the lane nodes, \ie, \textit{predecessor}, \textit{successor}, \textit{left neighbour} and \textit{right neighbour}.
For any lane node $A$, its predecessor and successor are defined as the neighbouring lane nodes that can travel to $A$ or from $A$ respectively. 
Note that one can reach the first lane node of a lane $l_{A}$ from the last lane node of lane $l_{B}$ if $l_{B}$ is the predecessor of $l_{A}$.
Left and right neighbours are defined as the spatially closest lane node measured by $\ell_2$ distance on the left and  on the right neighbouring lane respectively.
We denote the lane nodes with $V \in \mathbb{R}^{N \times 2}$, where $N$ is the number of lane nodes and the $i$-th row of $V$ is the BEV coordinates of the $i$-th node.
We represent the connectivity with $4$ adjacency matrices $\{A_i\}_{ i \in \{\text{pre}, \text{suc}, \text{left}, \text{right}\} }$, with $A_i \in \mathbb{R}^{N \times N}$.
We denote $A_{i, jk}$, as the element in the $j$-th row and $k$-th column of $A_i$. Then  $A_{i, jk} = 1$ if node $k$ is an $i$-type neighbor of node $j$. 

\subsubsection{LaneConv Operator:}
A natural operator to handle  lane graphs is the graph convolution \cite{shuman2013emerging}.
The most widely used graph convolution operator \cite{kipf2016semi} is defined as $Y = LXW$, where $X \in \mathbb{R}^{N \times F}$ is the node feature, $W \in \mathbb{R}^{F \times O}$ is the weight matrix, and $Y \in \mathbb{R}^{N \times O}$ is the output. The graph Laplacian matrix $L \in \mathbb{R}^{N \times N}$ takes the form $L = D^{-1/2}(I + A)D^{-1/2}$, where $I$, $A$ and $D$ are the identity, adjacency and degree matrices respectively. $I$ and $A$ account for self connection and connections between different nodes. All connections share the same weight $W$, and the degree matrix $D$ is used to normalize the output.
However, this vanilla graph convolution is inefficient in our case due to the following reasons. 
First, it is not clear what kind of node feature will preserve the information in the lane graphs.
Second, a single graph Laplacian can not capture the connection type, \ie, losing the directional information carried by the connection type.
Third, it is not straightforward to handle long range dependencies, \eg, akin dilated convolution, within this form of graph convolution.
Motivated by these challenges, we introduce our novel specially designed operator for lane graphs, called \textit{LaneConv}.

\paragraph{Node Feature:}
We first define the input feature of the lane nodes. Each lane node corresponds to a straight line segment of a centerline. To encode  all the lane node information, we need to take into account both the shape (size and orientation) and the location (the coordinates of the center) of the corresponding line segment. We parameterize the node feature as follows,
\begin{equation}\label{eqn:node_feat}
\mathbf{x}_i = \text{MLP}_\text{shape} \left( \mathbf{v}_i^{\text{end}} - \mathbf{v}_i^{\text{start}} \right)
+ \text{MLP}_{\text{loc}}\left(\textbf{v}_i\right),
\end{equation}
where $\text{MLP}$ indicates a multi-layer perceptron and the two subscripts refer to shape and location, respectively. 
$\textbf{v}_i$ is the location of the $i$-th lane node, \ie, the center between two end points, 
$\mathbf{v}_i^{\text{start}}$ and $\mathbf{v}_i^{\text{end}}$ are the BEV coordinates of the node $i$'s starting and ending points, and 
$\mathbf{x}_i$ is the $i$-th row of the node feature matrix $X$, denoting the input feature of the $i$-th lane node.

\paragraph{LaneConv:}
The node feature above only captures the local information of a line segment.
To aggregate the topology information of the lane graph at a larger scale,
we design the following LaneConv operator
\begin{equation}\label{eqn:laneconv}
Y = X W_0 + \sum_{i \in \{ \text{pre}, \text{suc}, \text{left}, \text{right} \}} {A_{i} X W_{i}},
\end{equation}
where $A_{i}$ and $W_i$ are the adjacency  and the weight matrices corresponding to the $i$-th connection type respectively. 
Since we order the lane nodes from the start to the end of the lane, $A_{\text{suc}}$ and $A_{\text{pre}}$ are matrices obtained by shifting the identity matrix one step towards upper right (non-zero superdiagonal) and lower left (non-zero subdiagonal). 
$A_{\text{suc}}$ and $A_{\text{pre}}$ can propagate information from the forward and backward neighbours whereas $A_{\text{left}}$ and $A_{\text{right}}$ allow information to flow from the cross-lane neighbours.
It is not hard to see that our LaneConv builds on top of the general graph convolution and encodes more geometric (\eg, connection type/direction) information.
As  shown in our experiments this  improves over the vanilla graph convolution.

\paragraph{Dilated LaneConv:}
Since motion forecasting models usually predict the future trajectories of actors with a time horizon of several seconds, actors with high speed could have moved a long distance.
Therefore, the model needs to capture the long range dependency along the lane direction for accurate prediction.
In regular grid graphs, a dilated convolution operator \cite{yu2015multi} can effectively capture the long range dependency by enlarging the receptive field.
Inspired by this operator, we propose the \textit{dilated LaneConv} operator to achieve a similar goal for irregular  graphs. 

In particular, the $k$-dilation LaneConv operator is defined as follows, 
\begin{equation}\label{eqn:dilated_laneconv}
Y = XW_0 + A_{\text{pre}}^k X W_{\text{pre},k} + A_{\text{suc}}^k X W_{\text{suc},k},
\end{equation}
where $A_{\text{pre}}^k$ is the $k$-th matrix power of $A_{\text{pre}}$. 
This  allows us to directly propagate information along the lane for $k$ steps, with $k$ a hyperparameter.
Since $A_{\text{pre}}^k$ is highly sparse, one can efficiently compute it using sparse matrix multiplication.
Note that the dilated LaneConv is only used for predecessor and successor, as the long range dependency is mostly along the lane direction.

\subsubsection{LaneGCN:}\label{sec:LGN}
\begin{figure*}[t]                                                                                                                           
	\begin{center}
		\includegraphics[width=0.85\linewidth]{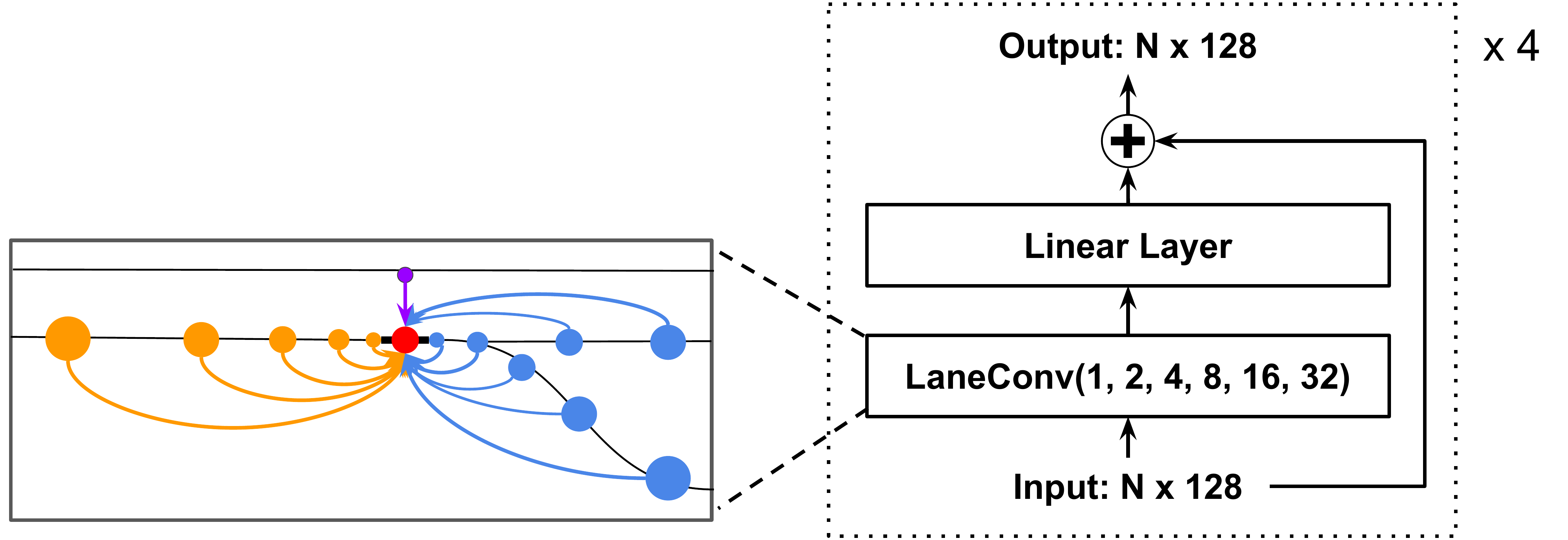}
	\end{center}
	\caption{\textbf{LaneGCN architecture}. Our LaneGCN is a stack of 4 multi-scale LaneConv residual blocks, each of which consists of a LaneConv(1,2,4,8,16,32) and a linear layer with a residual connection \cite{residual}. All layers have 128 feature channels.
	}
	\label{fig:LGN}
\end{figure*}

Based on the dilated LaneConv, we further propose a multi-scale LaneConv operator and use it to build our LaneGCN.
Combining Eq. (\ref{eqn:laneconv}) and (\ref{eqn:dilated_laneconv}) with multiple dilations, we get a multi-scale LaneConv operator with $C$ dilation sizes as follows
\begin{equation}\label{eqn:dilated_laneconv_final}
Y = XW_0 + \sum_{i \in \{ \text{left}, \text{right} \}} {A_{i} X W_{i}}
+ \sum_{c=1}^{C} {\left( A_{\text{pre}}^{k_{c}} X W_{\text{pre},k_{c}} + A_{\text{suc}}^{k_{c}} X W_{\text{suc},k_{c}} \right)},
\end{equation}
where $k_c$ is the $c$-th dilation size. We denote  $\text{LaneConv}(k_1, \cdots, k_C)$ this multi-scale layer. 
The architecture of LaneGCN is shown in Fig. \ref{fig:LGN}. The network is composed of $4$ LaneConv residual \cite{residual} blocks, which are the stack of a LaneConv(1, 2, 4, 8, 16, 32) and a linear layer, as well as a shortcut. All layers have 128 feature channels. Layer normalization \cite{layernorm} and ReLU \cite{glorot2011deep} are used after each LaneConv and linear layer.

\subsection{FusionNet}\label{sec:fusion}
In this section we propose a network to  fuse the information of the actor  and lane nodes given by ActorNet and MapNet, respectively. 
The behaviour of an actor strongly depends on its context, \ie, other actors and the map. 
Although the interactions between actors has been explored by previous work, the interactions between the actors and the map, and  map conditioned interactions between actors have received much less attention. 
In our model, we use spatial attention and LaneGCN to capture  a complete set of actor-map interactions (see Fig. \ref{fig:overall}).

We build a stack of four fusion modules to capture all information flows between actors and lane nodes, \ie, actors to lanes (A2L), lanes to lanes (L2L), lanes to actors (L2A) and actors to actors (A2A). 
Intuitively, A2L introduces real-time traffic information to lane nodes, such as blockage or usage of the lanes. 
L2L updates lane node features by propagating the traffic information over the lane graph. 
L2A fuses updated map features with real-time traffic information back to the actors. 
A2A handles the interactions between actors and produces the output actor features, which are then used by the prediction header for motion forecasting.

We implement L2L using another LaneGCN, which has the same architecture as the one used in our MapNet (see Section \ref{sec:LGN}). In the following we describe the other three modules in detail.
We exploit a spatial attention layer \cite{vaswani2017attention} for A2L, L2A and A2A. The attention layer applies to each of the three modules in the same way.
Taking A2L as an example, given an actor node $i$, we aggregate the features from its context lane nodes $j$ as follows
\begin{equation}\label{equ:attention}
\mathbf{y}_i = \mathbf{x}_i W_0 + \sum_j \phi ( \text{concat} (\mathbf{x}_i, \Delta_{i,j}, \mathbf{x}_j) W_1) W_2,
\end{equation}
with $\mathbf{x}_i$  the feature of the $i$-th node, $W$ a weight matrix, $\phi$ the composition of layer normalization and ReLU, and $\Delta_{ij} = \text{MLP}(\mathbf{v}_j - \mathbf{v}_i)$, where $\mathbf{v}$ denotes the node location.
The context nodes are defined to be the lane nodes whose $\ell_2$ distance from the actor node $i$ is smaller than a threshold. The thresholds for A2L, L2A and A2A are set to 7, 6, and 100 meters respectively.
Each of A2L, L2A and A2A has two residual blocks, which consist of a stack of the proposed attention layer and a linear layer, as well as a residual connection. All layers have 128 output feature channels.

\subsection{Prediction Header}\label{sec:header}
Taking the after-fusion actor features as input, a multi-modal prediction header outputs the final motion forecasting.
For each actor, it predicts $K$ possible future trajectories and their confidence scores. 
The header has two branches,  a regression branch to predict the trajectory of each mode and a classification branch to predict the confidence score of each mode.
For the $m$-th actor, we apply a residual block and a linear layer in the regression branch to regress the $K$ sequences of BEV coordinates:
\begin{equation}
O_{m, \text{reg}} = \{ (\mathbf{p}_{m,1}^k, \mathbf{p}_{m,2}^k, ..., \mathbf{p}_{m,T}^k) \}_{k \in [0, K-1]}
\end{equation}
where $\mathbf{p}_{m,i}^k$ is the predicted $m$-th actor's BEV coordinates of the $k$-th mode at the $i$-th time step.
For the classification branch, we apply an MLP to $\mathbf{p}_{m,T}^k - \mathbf{p}_{m,0}$ to get $K$ distance embeddings. 
We then concatenate each distance embedding with the actor feature, apply a residual block and a linear layer to output $K$ confidence scores, $O_{m, \text{cls}} = (c_{m,0}, c_{m,1}, ..., c_{m,K-1})$.

\subsection{Learning}
As all the modules are differentiable, we can train the model in an end-to-end way. 
We use the sum of   classification and regression losses to train the model
\begin{equation}
L = L_{\text{cls}} + \alpha L_{\text{reg}},
\end{equation}
where $\alpha=1.0$. 
Given $K$ predicted trajectories of an actor, we find a positive trajectory $\hat{k}$ that has the minimum final displacement error, \ie, the Euclidean distance between the predicted and ground truth locations at the final time step.

For classification, we use the max-margin loss:
\begin{equation}
L_{\text{cls}} = \frac{1}{M(K-1)}\sum_{m=1}^{M} \sum_{k\neq \hat{k}} \max( 0, c_{m,k} + \epsilon - c_{m,\hat{k}})
\end{equation}
where $\epsilon$ is the margin and $M$ is the total number of actors.
For regression, we apply the smooth $\ell1$ loss on all predicted time steps:
\begin{equation}
L_{\text{reg}} = \frac{1}{MT} \sum_{m=1}^{M} \sum_{t=1}^{T} \text{reg}(\mathbf{p}_{m,t}^{\hat{k}} - \mathbf{p}_{m,t}^*)           
\end{equation}
where $\mathbf{p}_t^*$ is the ground truth BEV coordinates at time step $t$, $\text{reg}(\mathbf{x})=\sum_id(x_i)$, $x_i$ is the $i$-th element of $\mathbf{x}$, and $d(x_i)$ is the smooth $\ell1$ loss defined as
\begin{equation}
d(x_i) = 
\begin{cases}
0.5 x_i^2& \text{if } \lVert x_i \rVert < 1\\
\lVert x_i \rVert - 0.5& \text{otherwise},
\end{cases}
\end{equation}
where $\lVert x_i \rVert$ denotes the $\ell_1$ norm of $x_i$. 

%% file: experiments.tex
\section{Experimental Evaluation}\label{sec:exp}
We evaluate our model on the large scale Argoverse \cite{chang2019argoverse} motion forecasting benchmark, which is publicly available and provides  vectorized map data.
We first compare our model with the state-of-the-art and show significant improvements in all metrics. We then conduct ablation studies on the architecture and LaneConv operators, and show the advantage of our model design choices.  Finally, we show qualitative results and discuss  future directions.

\subsection{Experimental Settings}
\subsubsection{Dataset:} Argoverse \cite{chang2019argoverse} is a motion forecasting benchmark with  over 30K scenarios collected in Pittsburgh and Miami. Each scenario is a sequence of frames sampled at 10 HZ. Each sequence has an interesting object called ``agent'', and the task is to predict the future locations of agents in a 3 seconds future horizon. The sequences are split into  training, validation and test sets, which have 205942, 39472 and 78143 sequences respectively. These splits  have no geographical overlap. For the training and validation sets, each sequence lasts for 5 seconds. The first two seconds are used as input data and the other 3 seconds are used as ground truth for models to predict. For the test set, only the first 2 seconds are provided. Each frame is given as the centroid coordinates of all objects in the scene. The actor data is a trajectory of 20 time steps. The map data is a set of lane centerlines and their connectivity. We use both actor and map data in the way described in Sections \ref{sec:actor} and \ref{sec:construct_graph}, without any other preprocessing step. We did not use the other map data such as the rasterized drivable area map and ground height map provided with the benchmark. 

\subsubsection{Metrics:} We employ two extensively used motion forecasting metrics,  {\it Average Displacement Error} (ADE) is defined as the $\ell_2$ distance between the predicted  and ground truth locations, averaged over all steps.  {\it Final Displacement Error} (FDE) is defined as the $\ell_2$ distance between the predicted  and ground truth locations at the last step in the predicted horizon. 
As motion forecasting is by nature multi-modal, Argoverse uses the minimum ADE (minADE) and minimum FDE (minFDE) of the top K predictions as the metrics. When K=1, minADE and minFDE are equal to the deterministic ADE and FDE. Argoverse benchmark allows up to 6 predictions, and the online server ranks the entries with minFDE with K=6. We use minADE and minFDE for K=1 and K=6 as the main metrics. When comparing our model with top entries on the leaderboard, we also show \textit{Miss Rate} (MR), which is the ratio of predictions (the best mode) whose final location is more than 2.0 meters away from the ground truth.

\subsubsection{Implementation Details:} \label{sec:implementation}
We use all actors and lanes whose distance from the agent is smaller than 100 meters as the input. The coordinate system in our model is the BEV centered at the agent location at $t=0$.  We use the orientation from the agent location at $t=-1$ to the agent location at $t=0$ as the positive x axis.
We train the model on 4 TITAN-X GPUs using a batch size of 128 with the Adam \cite{adam} optimizer with an initial learning rate of $1\times 10^{-3}$, which is decayed to $1\times 10^{-4}$ at 32 epochs. The training process finishes at 36 epochs and takes about 11.5 hours. All our results are based on the same model, whose architecture and hyper-parameters are described in Section \ref{sec:model}. 

\subsection{Results}
\subsubsection{Comparison with the state-of-the-art:}

\begin{table*}[t]
	\begin{center}
		\caption{Results on Argoverse motion forecasting benchmark (test set)} \label{tab:test}
		\begin{tabular}{l||ccc|ccc}
			\hline
			\multirow{2}{*}{Model} & \multicolumn{3}{c|}{K=1} & \multicolumn{3}{c}{K=6}\\
			& minADE & minFDE & MR & minADE & \textbf{minFDE} & MR\\
			\hline
			Argoverse Baseline \cite{chang2019argoverse} & 2.96 & 6.81 & 0.81 & 2.34 & 5.44 & 0.69\\
			Argoverse Baseline (NN) \cite{chang2019argoverse} & 3.45 & 7.88 & 0.87 & 1.71 & 3.29 & 0.54\\
			\hline
			Holmes (\textit{7th}) \cite{huang2019diversity}  & 2.91 & 6.54 & 0.82 & 1.38 & 2.66 & 0.42\\
			cxx (\textit{3rd}) \cite{argoai_challenge} & 1.91 & 4.31 & 0.66 & 0.99 & 1.71 & 0.19\\ 
			uulm-mrm (\textit{2nd}) \cite{Cui2018MultimodalTP,djuric2018motion} & 1.90 & 4.19 & 0.63 & 0.94 & 1.55 & 0.22\\
			Jean (\textit{1st}) \cite{argoai_challenge,jean2019} & 1.86 & 4.18 & 0.63 & 0.93 & 1.49 & 0.19\\
			\hline
			Our Model & \textbf{1.71} & \textbf{3.78} & \textbf{0.59} & \textbf{0.87} & \textbf{1.36} & \textbf{0.16}\\
			\hline
		\end{tabular}
	\end{center}
\end{table*}

 We  compare our model with four top entries and two official baselines on the Argoverse motion forecasting leaderboard. We submit our result at the time of ECCV submission (2020/03/15). The metrics are minADE, minFDE and MR for K=1 and K=6, and the leaderboard is ranked by minFDE for K=6. As shown in Table \ref{tab:test}, our model significantly outperforms all other models in all metrics.
Among the compared methods, uulm-mrm encodes the input data using a rasterization approach \cite{Cui2018MultimodalTP,djuric2018motion}. They represent actor states, lanes and the drivable area with a synthesized image, which is then processed by a 2D CNN. In this approach, map topology and actor-map interactions are both implicitly learned by 2D convolution. In contrast, our model explicitly learns structured map features and performs actor-map fusion. Jean and cxx encode actors and lanes with 1D CNN and/or LSTM, and use attention \cite{vaswani2017attention} to fuse the features. In their models, lanes are encoded independently so the global map topology is not captured. Moreover, there is no actor to lane and lane to lane fusion. In contrast, our model learns the lane features using the LaneConv, which captures the multi-scale topology of the lane graph. 

\subsubsection{Importance of each module:}
\begin{table*}[t]
	\begin{center}
		\caption{Ablation study results of modules} \label{tab:module_ablation}
		\begin{tabular}{cc|cccc||cc|cc}
			\hline
			\multicolumn{2}{c|}{Backbone} & \multicolumn{4}{c||}{FusionNet} & \multicolumn{2}{c|}{K=1} & \multicolumn{2}{c}{K=6}\\
			ActorNet & MapNet & L2A & A2L & L2L & A2A & minADE & minFDE & minADE & minFDE\\
			\hline
			\checkmark & & & & & & 1.90 & 4.38 & 0.91 & 1.66\\
			\checkmark & & & & & \checkmark & 1.58 & 3.61 & 0.79 & 1.29\\
			\hline
			\checkmark & \checkmark& \checkmark& & & & 1.55 & 3.52 & 0.76 & 1.23 \\
			\checkmark & \checkmark& \checkmark& \checkmark & \checkmark & & 1.39 & 3.05 & 0.72 & 1.10 \\
			\checkmark & \checkmark& \checkmark& \checkmark & \checkmark & \checkmark & \textbf{1.35} & \textbf{2.97}& \textbf{0.71} & \textbf{1.08} \\
			\hline
		\end{tabular}
	\end{center}
\end{table*}

In Table \ref{tab:module_ablation}, we show the results of using ActorNet as the baseline and progressively adding more modules. Three observations can be drawn from the results. First, all modules improve the performance of the model,  demonstrating the effectiveness of both LaneGCN and our overall architecture. Second, the information flow from actors to maps brings useful traffic information which benefits the motion forecasting performance, as the incorporation of A2L and L2L significantly outperforms L2A only. 
Third, A2L, L2L and L2A also facilitates the interaction between actors, which can be seen from the smaller gain of adding A2A to this combination (from 4th row to 5th row) compared to adding A2A to ActorNet alone (from 1st row to 2nd row). Intuitively, the information of different actors is propagated over the lane graph and leads to effective map conditioned interactions.

\subsubsection{Lane Graph Operators:}\label{sec:ablation_graph}
\begin{table*}[t]
	\begin{center}
		\caption{Ablation study results of lane graph operators} \label{tab:graph_ablation}
		\begin{tabular}{cccc||cc|cc}
			\hline
			\multicolumn{4}{c||}{Component} & \multicolumn{2}{c|}{K=1} & \multicolumn{2}{c}{K=6}\\
			GraphConv & Residual & Multi-Type & Dilate & minADE & minFDE & minADE & minFDE\\
			\hline
			\checkmark & & & & 1.72 & 3.93 & 0.82 & 1.41\\
			\checkmark & & \checkmark & & 1.59 & 3.59 & 0.77 & 1.24\\
			\checkmark & & & \checkmark & 1.46 & 3.29 & 0.74 & 1.16\\
			\hline
			\checkmark & \checkmark & & & 1.53 & 3.48 & 0.79 & 1.33\\
			\checkmark & \checkmark & \checkmark & & 1.48 & 3.33 & 0.74 & 1.19\\
			\checkmark & \checkmark & & \checkmark & 1.41 & 3.12 & 0.73 & 1.14\\
			\hline
			\checkmark & \checkmark & \checkmark & \checkmark & \textbf{1.39} & \textbf{3.05} & \textbf{0.72} & \textbf{1.10}\\
			\hline
		\end{tabular}
	\end{center}
\end{table*}

\begin{figure*}[t]                                                                                                                           
	\begin{center}
		\includegraphics[width=1.0\linewidth]{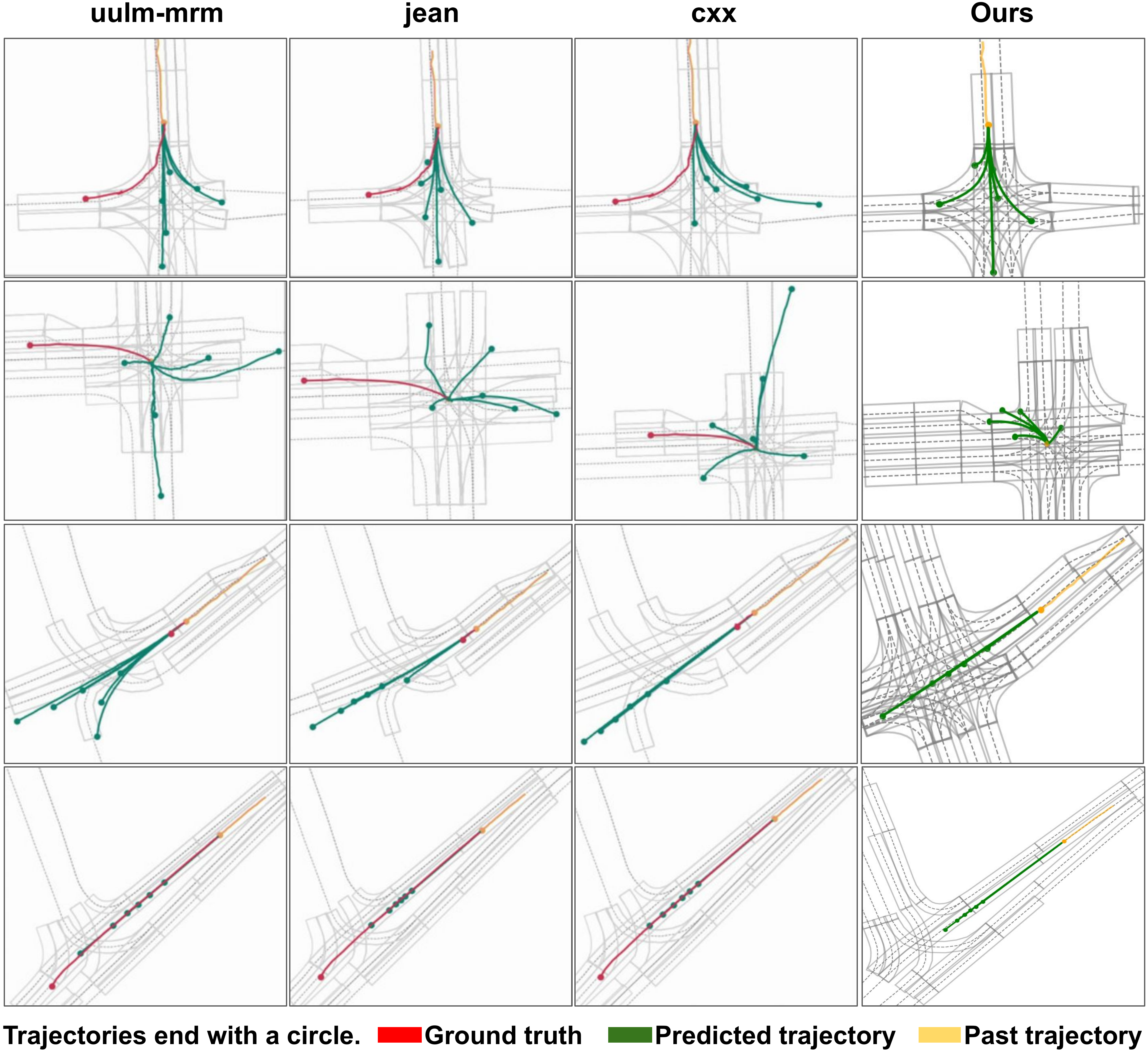}
	\end{center}
	\caption{\textbf{Qualitative results on hard cases.} From top to bottom, these hard cases involve missing the right turn mode, lacking history information, extreme deceleration and acceleration, respectively. See the text for more information.}
	\label{fig:hard}
\end{figure*}

In Table \ref{tab:graph_ablation}, we show the results of the ablation study on lane graph operators. The baseline model uses the combination of A2L, L2L and L2A. We start from the vanilla graph convolution (GraphConv), and evaluate the effect of adding each component of the LaneConv block (see Figure \ref{fig:LGN}), including the residual block, multi-type connections and dilation. 
The last row is the LaneConv used in our model (fourth row of Table \ref{tab:module_ablation}). All these components significantly improve the performance.
The residual block only adds about $7\%$ parameters, but effectively facilitates the training. Both multi-type connections and dilation significantly boost the performance, demonstrating the clear advantage of LaneConv over  vanilla graph convolution.

\subsubsection{Qualitative Results:}\label{sec:hard}
In Fig. \ref{fig:hard}, we compare qualitatively our model to other methods on 4 hard cases. The results of other models are adapted from the slides of Argoverse motion forecasting competition \cite{argoai_challenge}. As the examples are from the test set and we have no access to the labels, in our results we did not show the ground truth trajectory.
The first row shows a case where the baselines miss the mode. 
While the other methods fail to capture the right turn prediction, our model produces a mode which nicely follows the right turn centerline. The second row shows a case  where the agent  is waiting  to perform an unprotected left turn for the first 2 seconds. Due to the lack of actor motion  history, maps are important for the model to produce reasonable trajectories. 
The other models produce divergent trajectories, some of which are non-traffic-rule compliant. In contrast, our model produces reasonable trajectories following the lane topology. The third row shows a case of a car decelerating and coming to a stop at the intersection. Our model produces a mode with more deceleration then the baselines and all the modes reasonably follow the lane. The fourth row shows a case of extreme acceleration. None of the models captures this case well, possibly because there is not enough information to make this prediction.

Overall, these results suggest that  LaneGCN effectively learns  structured map representations, which are used by the model to predict realistic trajectories. One potential way to improve our model is to incorporate more map information into the lane graph. Currently our model uses the centerlines and their connectivity. Other map information, such as traffic lights and traffic signs, provides useful information for motion forecasting, which is well illustrated by the second and third cases in Fig. \ref{fig:hard}. To account for new map data, our model can be easily extended by introducing new nodes and connections. We will explore this direction in future work.

%% file: conclusion.tex
\section{Conclusion}\label{sec:conclusion}
In this paper, we propose a novel motion forecasting model to learn lane graph representations and perform a complete set of actor-map interactions.
Instead of using a rasterized map as input, we construct a lane graph from vectorized map data and propose the LaneGCN to extract map topology features. We use spatial attention and the LaneGCN to fuse the information of both actors and lanes. We conduct experiments on the large scale Argoverse motion forecasting benchmark. Our model significantly outperforms the state-of-the-art.
In the future  we plan to explore the incorporation of other map data.

%% file: acknowledgement.tex
\section*{Acknowledgement}
We want to thank Wenyuan Zeng and Cole Gulino for their helpful comments on the paper.

%% file: appendix.tex
\section*{Appendix}
\begin{figure*}[t]
	\begin{center}
		\includegraphics[width=1.0\linewidth]{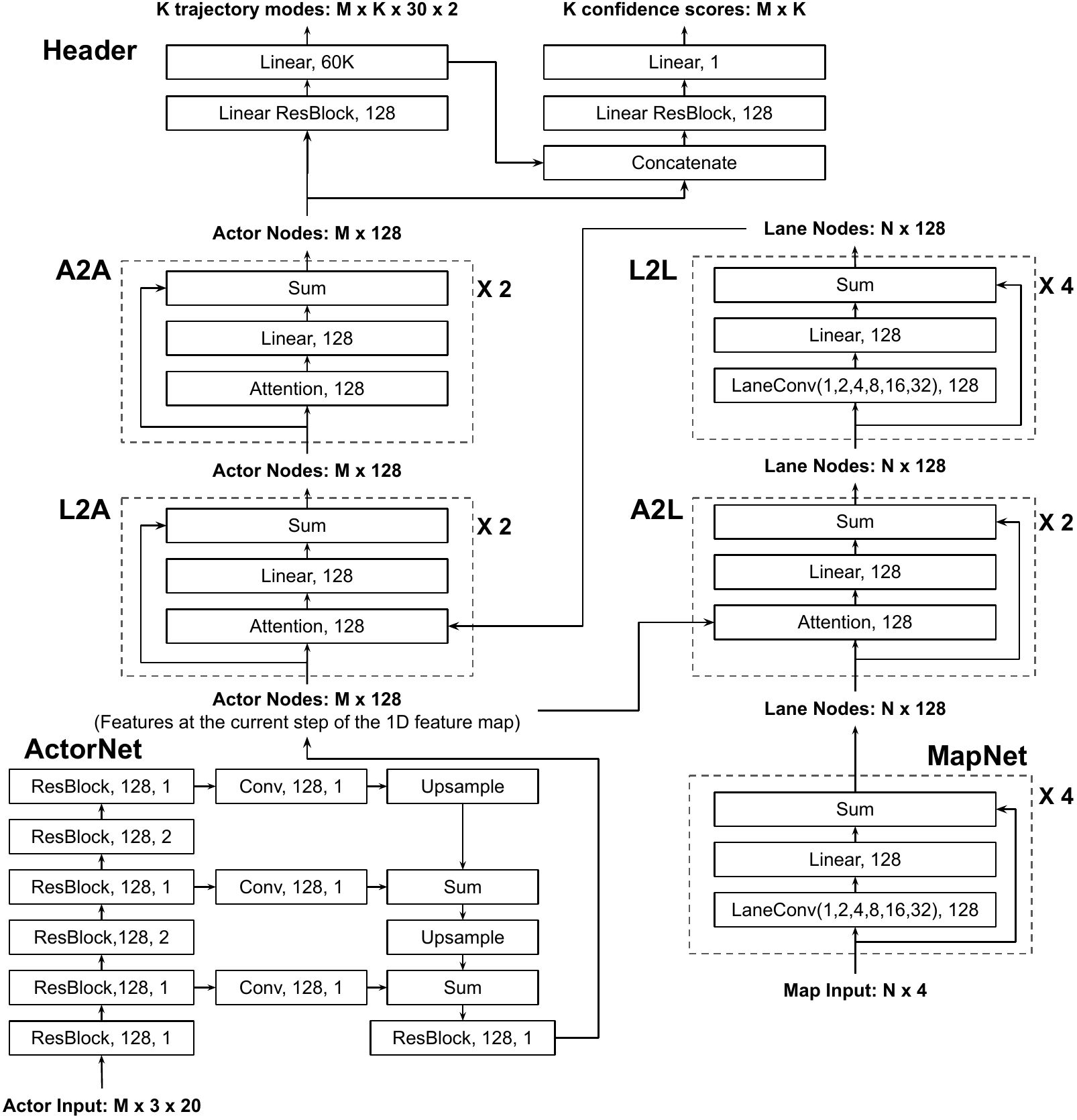}
	\end{center}
	\caption{\textbf{Detailed Architecture}.
			Learnable blocks are named in the form of \textit{layer type, output channels, stride} (no stride for linear layers). 
			 Upsample, Sum and Concatenate denote bilinear upsampling, element-wise summation and feature concatenation layers, respectively. LaneConv and Attention are described by Equation (\ref{eqn:dilated_laneconv_final}) and (\ref{equ:attention}).
	}
	\label{fig:detail}
\end{figure*}

We show the detailed architecture of our model in Figure \ref{fig:detail}.
Our model is composed of 4 modules, ActorNet, MapNet, Actor-Map Fusion Cycle, and the Prediction Header. ActorNet extracts temporal features with a 1D CNN and merges the multi-scale features with a Feature Pyramid Network \cite{Lin2016FeaturePN}. MapNet is a Lane Graph Network (LGN), which extracts lane topology features with the proposed LaneConv operators. The LGN is a stack of 4 multi-scale LaneConv residual blocks. Actor-map fusion cycle is a stack of 4 fusion networks, including actor-to-lane (A2L), lane-to-lane (L2L), lane-to-actor (L2A), actor-to-actor (A2A). A2L, L2A and A2A are a stack of 2 attention residual blocks. L2L is another LGN. Finally, the updated actor features are used by the prediction header to produce the multi-modal trajectories and their confidence scores.